%% file: main.tex
\documentclass[runningheads]{llncs}

\usepackage{accv}

\usepackage{accvabbrv}

\input{preamble}
\usepackage{graphicx}
\usepackage{booktabs}
\usepackage{colortbl}
\usepackage{arydshln}
\usepackage{multirow}
\usepackage{tabu}
\usepackage{makecell}
\usepackage{wrapfig}
\usepackage{float}

\usepackage[accsupp]{axessibility}  

\usepackage{hyperref}

\usepackage{orcidlink}

\begin{document}

\title{MTV: Revisiting Multi-Task Visual Representation Learning}


\author{Shangzhe Di\inst{1,2}\orcidlink{0009-0005-6977-4332} \and
Zhonghua Zhai\inst{2}\orcidlink{0009-0001-4878-1892} \and
Weidi Xie\inst{1}\orcidlink{0009-0002-8609-6826}}

\authorrunning{Shangzhe Di, Zhonghua Zhai, Weidi Xie}

\institute{SAI, Shanghai Jiao Tong University \and
ByteDance Seed \\
\url{https://github.com/Becomebright/MTV}
}

\maketitle

\input{sections/0_abstract}
\input{sections/1_intro}
\input{sections/2_related}
\input{sections/3_method}
\input{sections/4_experiments}
\input{sections/5_conclusion}

%
%
\bibliographystyle{splncs04}
\bibliography{main}

\input{sections/X_suppl}

\end{document}

%% file: preamble.tex
\usepackage[dvipsnames]{xcolor}

\definecolor{mygreen}{RGB}{29, 177, 0}
\definecolor{darkgreen}{rgb}{0.0, 0.5, 0.0}

\newcommand{\plusvalue}[1]{\hspace{0.1em}\textcolor{darkgreen}{{\scriptsize{(#1)}}}}
\newcommand{\Plusvalue}[1]{\textcolor{darkgreen}{\textbf{#1}}}
\newcommand{\minusvalue}[1]{\hspace{0.1em}\textcolor{red}{{\scriptsize{(#1)}}}}

\newcommand{\addvalue}[1]{\hspace{0.1em}\textcolor{black!60}{{\scriptsize{(#1)}}}}

\def\ourmethod{MTV}

%% file: sections/0_abstract.tex
\begin{abstract}
  Current visual representation learning remains bifurcated: vision-language models (e.g., CLIP) excel at global semantic alignment but lack spatial precision, while self-supervised methods (e.g., MAE, DINO) capture intricate local structures yet struggle with high-level semantic context. We argue that these paradigms are fundamentally complementary and can be integrated into a principled multi-task framework, further enhanced by dense spatial supervision. We introduce \textbf{\ourmethod}, a \underline{m}ulti-\underline{t}ask \underline{v}isual pretraining framework that jointly optimizes a shared backbone across vision-language contrastive, self-supervised, and dense spatial objectives. To mitigate the need for manual annotations, we leverage high-capacity ``expert'' models—such as Depth Anything V2 and OWLv2—to synthesize dense, structured pseudo-labels at scale. Beyond the framework, we provide a systematic investigation into the mechanics of multi-task visual learning, analyzing: (i) the individual and (ii) marginal gain of each objective, (iii) task synergies versus interference, and (iv) scaling behavior across varying data and model scales. Our results demonstrate that \ourmethod~achieves ``best-of-both-worlds'' performance, significantly enhancing fine-grained spatial reasoning without compromising global semantic understanding. Our findings suggest that multi-task learning, fueled by high-quality pseudo-supervision, is a scalable path toward more general visual encoders.
  \keywords{Visual representation learning \and Multi-task training}
\end{abstract}

%% file: sections/1_intro.tex
\section{Introduction}
\label{sec:intro}

\input{images/teaser}

Recent advances in foundation models have reshaped visual representation learning by scaling pretraining on web-scale data.
Currently, two paradigms prevail:
\textit{Vision–language contrastive learning}~\cite{clip,align,siglip} aligns global image and text embeddings via instance-level discrimination, enabling open-vocabulary recognition and zero-shot transfer. \textit{Self-supervised learning}, such as masked image modeling~\cite{mae,ibot}, contrastive instance discrimination~\cite{moco,simclr}, and student–teacher matching~\cite{dino,dinov2}, enforces invariance and regularities without manual labels.
Despite their success, these paradigms remain bifurcated, primarily focusing on either global semantics or local regularities while leaving fine-grained spatial and geometric reasoning as under-constrained emergent properties.
Given the inherent diversity of visual tasks—ranging from low-level geometry to high-level reasoning—learning truly universal visual representations that bridge these multifaceted requirements remains an open and fundamental challenge~\cite{probe3d,tong2024cambrian,mmvp,liu-etal-2025-data-language}.

In contrast, large language models (LLMs) scale effectively by absorbing multifaceted supervision—such as summarization, translation, and question answering—within a unified training interface~\cite{scaling_law,gpt3,gpt4}.
Admittedly, unlike natural language, visual tasks are inherently heterogeneous. While this diversity makes a single, universal objective more elusive in vision, we argue that visual pretraining can still embrace the LLM philosophy by simultaneously optimizing across a diverse suite of tasks to capture the full spectrum of information.

To bridge this gap, we introduce \textbf{\ourmethod}, a multi-task visual pretraining framework that jointly optimizes a shared backbone across vision-language contrastive, self-supervised, and dense spatial objectives, as shown in Fig.~\ref{fig:teaser}.
To overcome the scarcity of human-labeled data, we leverage high-capacity ``expert'' models—such as Depth Anything V2~\cite{depth_anythingv2} for monocular depth estimation and OWLv2~\cite{owlv2} for open-vocabulary grounding—to synthesize dense, structured pseudo-labels at scale.
By integrating these expert-guided targets, we enrich traditional objectives with explicit geometric and spatial priors.

Besides, we go beyond simple multi-tasking to provide a systematic investigation into the mechanics of multi-task visual learning.
We create a controlled setting to study how heterogeneous signals cooperate.
Specifically, we examine: (i) the individual and (ii) marginal gain of each objective, (iii) the dynamics of task synergies versus interference, and (iv) scaling behavior across varying data and model scales.
Empirically, \ourmethod~achieves ``best-of-both-worlds'' performance, yielding representations that are more general and data-efficient than any single objective alone.
For instance, our \texttt{\ourmethod-Base} model trained on 100M samples achieves 69.5\% ImageNet zero-shot accuracy—outperforming \texttt{CLIP-Base}~\cite{clip} trained on 400M samples—while delivering substantial gains across retrieval, depth estimation, segmentation, and correspondence benchmarks.

Taken together, these findings suggest a clear design principle: expanding conventional visual pretraining with complementary tasks—even when supervised by imperfect pseudo labels—shifts learning toward a unified, multi-granularity representation space.
This unification, analogous in spirit to the convergence of diverse language tasks for LLM, offers a scalable and data-efficient pathway for building versatile visual encoders capable of anchoring a wide range of perception and multimodal reasoning tasks.

%% file: images/teaser.tex
\begin{wrapfigure}[18]{R}{0.6\textwidth}
  \vspace{-25pt}
  \centerline{\includegraphics[width=.95\linewidth]{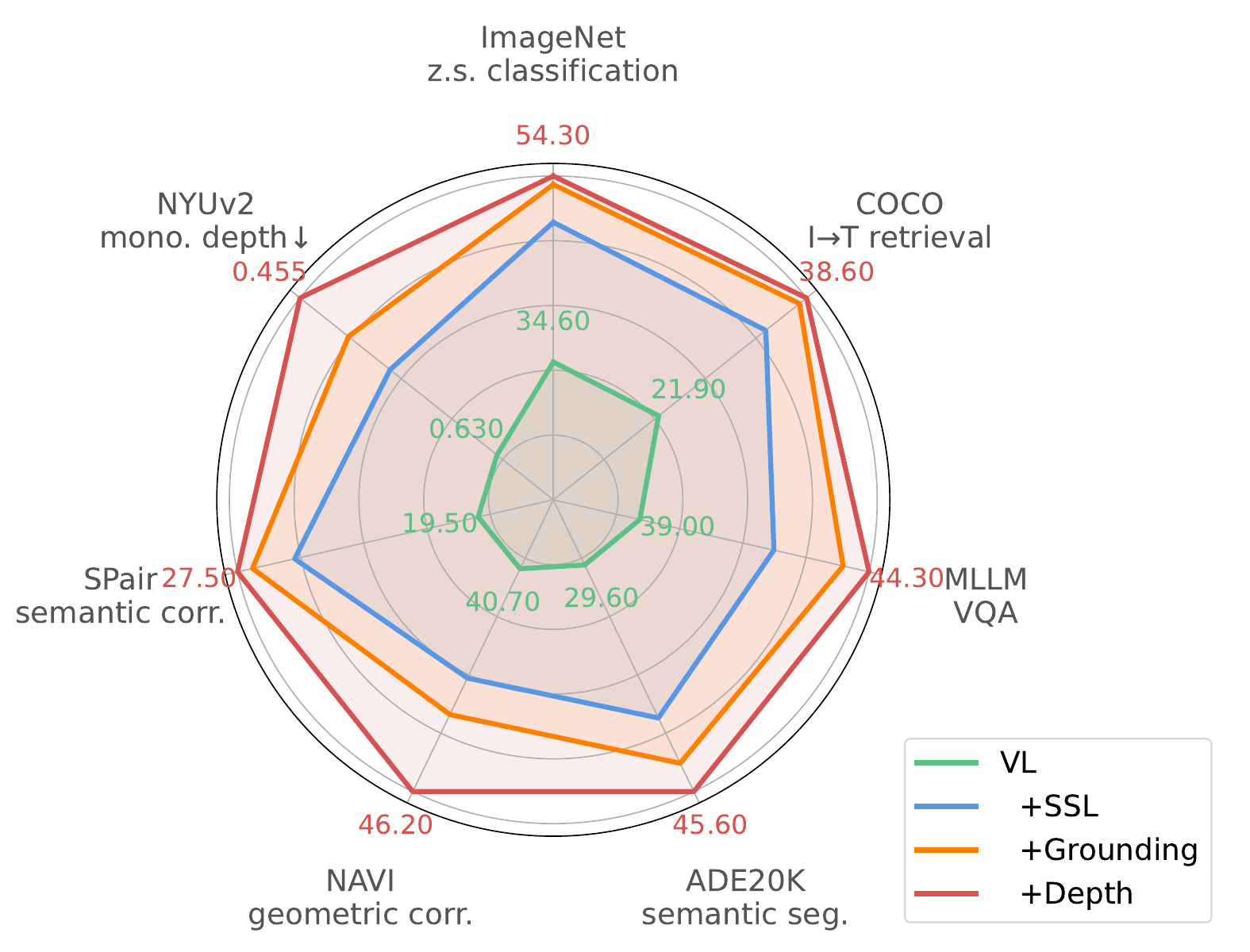}}
  \vspace{-5pt}
  \caption{
    \textbf{Multi-task supervision effects} on a \texttt{ViT-L} model with 10M training samples.
    We integrate \textcolor[HTML]{5DC086}{VL}, \textcolor[HTML]{5896E6}{SSL}, and pseudo-labeled \textcolor[HTML]{FF7F00}{grounding} and \textcolor[HTML]{D6534F}{depth} estimation into a unified training framework, which leads to strong and consistent performance gains across diverse tasks.
  }\label{fig:teaser}
  \vspace{-10pt}
\end{wrapfigure}

%% file: sections/2_related.tex
\section{Related Work}
\label{sec:related}

\noindent\textbf{Self-supervised Learning (SSL).}
Self-supervised visual pretraining aims to learn transferable representations without manual annotations.
Contrastive and distillation-based methods, such as DINO~\cite{dino}, iBOT~\cite{ibot}, and I-JEPA~\cite{ijepa}, encourage feature invariance across augmented views, whereas reconstruction-based approaches, like MAE~\cite{mae}, rely on masked image modeling.
Recent large-scale efforts such as DINOv2~\cite{dinov2} demonstrate that scaling data and model size leads to robust features for recognition, segmentation, and correspondence tasks.
However, these methods remain limited to unimodal visual signals and lack semantics from natural language or high-level reasoning~\cite{tong2024cambrian}.

\vspace{8pt}
\noindent\textbf{Vision-Language Contrastive Learning (VL).}
Weakly-supervised vision-language pretraining has emerged as a dominant paradigm for learning semantic-aligned visual encoders.
CLIP~\cite{clip} and ALIGN~\cite{align} show that large-scale vision-language contrastive learning yields powerful zero-shot transfer.
Subsequent works such as Florence~\cite{florence} and SigLIP~\cite{siglip} further improve alignment quality and scalability through better data curation and sigmoid-based objectives.
Despite their impressive semantic generalization, these models are optimized purely for global alignment and often underperform on spatially structured or geometric tasks such as depth and correspondence estimation~\cite{probe3d}.

\vspace{8pt}
\noindent\textbf{Self-Training (ST).}
Self-training refers to using a supervised model to generate pseudo labels on unlabeled data, followed by training a student model on these pseudo-labeled samples~\cite{xie2020self}.
This paradigm has been widely adopted to expand supervision beyond human-annotated datasets.
For VL contrastive learning, several approaches~\cite{capsfusion,veclip,LaCLIP,clips} generate synthetic captions or rewrite noisy web captions to enhance CLIP-style training.
However, textual descriptions alone cannot fully capture the rich visual content.
Beyond VL pretraining, self-training has also demonstrated remarkable success in other visual domains.
For instance, Depth Anything~\cite{depth_anything,depth_anythingv2} first trains a teacher model on high-quality synthetic depth data and then uses it to generate pseudo labels on real images for student training, leading to state-of-the-art monocular depth estimation.
Similarly, OWLv2~\cite{owlv2} employs a supervised open-vocabulary detector to produce billion-scale pseudo labels for scaling.
Overall, self-training has emerged as an effective strategy for scaling visual supervision by leveraging strong teacher models to generate reliable annotations at an unprecedented scale.
We build upon this trend by extending self-training to multi-task settings that combine VL and SSL objectives, enabling more general and scalable visual representation learning.

\vspace{8pt}
\noindent\textbf{Multi-task Visual Representation Learning.}
Multi-task learning optimizes shared representations to capture cross-domain structures and enhance generalization. 
Methods such as SLIP~\cite{slip}, SILC~\cite{silc}, and TIPS~\cite{tips} show that integrating VL and SSL objectives yields complementary supervision, improving both semantic alignment and spatial understanding.
In the context of ST, MuST~\cite{must} pioneered the idea that combining multiple self-training tasks can yield stronger representations than purely supervised or self-supervised learning.
We advance this trajectory by integrating VL, SSL, and ST under a unified framework.
Notably, while SigLIP2~\cite{siglip2} recently explored a similar combination of VL, SSL, and grounding-based ST~\cite{locca}, it lacks a systematic analysis of how task composition and scaling impact the learned feature space. Our work diverges by providing rigorous studies across task diversity and data/model scales, and crucially, by introducing depth supervision into the multi-task objective. We demonstrate that this geometric cue not only enhances dense spatial understanding but also unexpectedly bolsters global semantic alignment.


%% file: sections/3_method.tex
\section{Preliminary} \label{sec:formulation}

Our goal is to learn unified visual representations that support a broad spectrum of visual understanding tasks via large-scale, multi-source supervision. We begin by introducing the key components used throughout the paper, covering both the shared architecture and the heterogeneous annotations that supervise it.

\vspace{5pt} \noindent \textbf{Architecture.}
At the core of our framework is a shared ViT~\cite{vit} encoder $f_\theta$ that maps an image $v_i \in \mathbb{R}^{H \times W \times 3}$ to multi-layer patch features:
\begin{equation*}
  \{\mathbf{z}_i^{(1)}, \dots, \mathbf{z}_i^{(L)}\} = f_\theta(v_i), \quad \mathbf{z}_i^{(l)} \in \mathbb{R}^{N \times D},
\end{equation*}
where $L$ is the number of layers, $N$ the number of patches, and $D$ the feature dimension.
For brevity, we denote the final layer representation as $\mathbf{z}_i := \mathbf{z}_i^{(L)}$.
For tasks requiring a global image-level representation, we apply attentive pooling~\cite{siglip} to the final patch features: $\mathbf{v}_i = \text{Pool} ( \mathbf{z}_i ) \in \mathbb{R}^D$.

Simultaneously, a text encoder $g_\phi$ processes natural language inputs.
To enable cross-modal alignment, $g_\phi$ maps text $t_i$ into a joint latent space shared with the pooled visual features, producing the text embedding $\mathbf{t}_i = g_\phi(t_i) \in \mathbb{R}^D$.

\vspace{5pt}
\noindent \textbf{Training Data.}
We consider a pretraining corpus $\mathcal{D} = \{(v_i, a_i)\}$, where each image $v_i$ is paired with automatically acquired annotations $a_i$ that provide semantic, spatial, and geometric signals (Fig.~\ref{fig:framework}a):
(i) A \textbf{caption} $t_i$ capturing the global semantics of the image,
used for vision-language contrastive learning;
(ii) \textbf{Region–text pairs} $\{(b_{ij}, r_{ij})\}$
where $b_{ij}$ and $r_{ij}$ denote the $j$-th bounding box and its entity name.
These pairs are obtained by extracting salient entities using \texttt{RAM++}\cite{ram++},
and then prompting the open-vocabulary detector \texttt{OWLv2-Base}\cite{owlv2} to localize the referred regions;
(iii) A \textbf{relative depth map} $d_i \in \mathbb{R}^{H \times W}$ that encodes per-pixel depth, produced by the \texttt{Depth-Anything-V2-Large} model~\cite{depth_anythingv2}.
In the supplementary material, we also provide results using a smaller teacher model and show that our method is robust to teacher capacity.

These heterogeneous annotations, derived entirely from web data and off-the-shelf specialist models, provide a rich supervisory signal for studying the effectiveness of large-scale multi-task representation learning.

\section{Multi-task Visual Representation Learning}
\input{images/framework}

This section details \textbf{\ourmethod}, a principled \underline{m}ulti-\underline{t}ask \underline{v}isual representation learning framework.
As shown in Fig.~\ref{fig:framework}b, \ourmethod~optimizes a shared encoder across three complementary supervision paradigms: (i) vision-language contrastive learning for global semantic alignment, (ii) self-supervised learning to induce local structural invariance, and (iii) dense supervision to capture fine-grained spatial structures.
We elaborate on each supervisory component below.

\subsection{Global Semantic Supervision}

We adopt vision–language contrastive learning to provide global semantic supervision by aligning image and text embeddings at the instance level. This objective enables the model to learn high-level semantic representations from noisy, large-scale web data.

Given a batch of image–text pairs $\mathcal{B} = \{(v_i, t_i)\}_{i=1}^B$, the visual and textual embeddings $\mathbf{v}_i$ and $\mathbf{t}_i$ are obtained using the visual encoder and text encoder, as defined in Section~\ref{sec:formulation}.
We adopt the sigmoid-based objective from SigLIP~\cite{siglip}, which treats the alignment task as a dense pairwise binary classification problem:
\begin{equation}
  \mathcal{L}_{\text{VL}} = -\frac{1}{B} \sum_{i=1}^{B} \sum_{j=1}^{B} \log \frac{1}{1+e^{y_{i j}\left(-\tau \mathbf{v}_i \cdot \mathbf{t}_j+\beta\right)}},
  \label{equ:vl}
\end{equation}
where $y_{ij} = 1$ if $(v_i, t_j)$ is a matched pair and $-1$ otherwise. The logit scale $\tau$ and bias $\beta$ are learnable parameters initialized from $\log 10$ and $-10$, respectively.
Unlike the softmax-based contrastive loss~\cite{clip}, the sigmoid objective decouples the batch size from the normalization, leading to more stable gradients at scale.

To support large-scale distributed training, SigLIP originally employs a cyclic rotation-based protocol to avoid large collective communication. However, in our framework, we observe that the communication overhead of text embeddings is negligible relative to the computational cost of the shared multi-task backbone. Therefore, we utilize a more direct and implementation-efficient alternative: we synchronize textual embeddings across all devices using a differentiable \texttt{all\_gather}\footnote{\texttt{all\_gather} is a collective communication primitive that synchronizes and collects tensors from all devices. We use a differentiable version of it that supports gradient backpropagation through $\mathbf{t}_j$.} operation. This preserves the mathematical correctness of Eq.~\ref{equ:vl} while facilitating easier integration within a multi-task pipeline.

\subsection{Self-Supervised Objectives}

While VL supervision captures global semantics, it provides limited guidance for learning locally consistent features.
To complement it, we incorporate two self-supervised objectives within a teacher-student distillation framework: (i) local-to-global self-distillation~\cite{dino} and (ii) masked feature prediction~\cite{ibot}.

Following the architecture in DINOv2~\cite{dinov2} and TIPS~\cite{tips}, our student network comprises the visual encoder $f_\theta$ and separate MLP projection heads for each SSL task to mitigate gradient interference.
The teacher network is maintained as an Exponential Moving Average (EMA) of the student, providing stable targets throughout pretraining.

\vspace{5pt}
\noindent\textbf{Local-to-Global Self-distillation} encourages spatial consistency by aligning the representations of local and global views.
For each image $v_i$, the student processes $M=6$ local crops to produce their global embeddings $\{\mathbf{v}_{i,m}\}_{m=1}^M$, which are then projected into ``prototype scores'' $\{\mathbf{p}_{i,m}\}_{m=1}^M$ by the student MLP. Simultaneously, the teacher processes a larger, global crop to produce the target scores $\hat{\mathbf{p}}_{i}$.
The student is trained to match the teacher's distribution using a temperature-scaled KL divergence:
\begin{equation*}
  \mathcal{L}_\text{distill} = -\sum_i^B \sum_m^M
  \text{KL}\left(
    \text{softmax} (\frac{\hat{\mathbf{p}}_{i} - \mathbf{c}}{\tau_t} )
    \Big\|
  \text{softmax} (\frac{\mathbf{p}_{i,m}}{\tau_s}) \right),
\end{equation*}
where $\tau_t$ and $\tau_s$ are the teacher and student temperatures for score sharpening, and $\mathbf{c}$ is an EMA-updated centering variable that prevents collapsing to a trivial solution by encouraging a uniform distribution across prototypes.

\vspace{3pt}
\noindent \textbf{Masked Feature Prediction} focuses on capturing fine-grained spatial dependencies.
We randomly mask 50\% of the patches in $v_i$ and replace them with a learnable mask token.
The student encodes the masked input into prototype scores $\{\mathbf{p}_i^j\}$, while the teacher processes the original unmasked image to provide target scores $\{\hat{\mathbf{p}}_i^j\}$ for the corresponding masked locations $j$.
The student is trained to match the teacher’s distributions at these masked locations, thereby forced to model long-range spatial regularities from incomplete inputs:
\begin{equation*}
  \mathcal{L}_\text{mask} = -\sum_i^B \sum_{j \in \text{masked}}
  \text{KL}\left(
    \text{softmax} (\frac{\hat{\mathbf{p}}_i^j - \mathbf{c}}{\tau_t} )
    \Big\|
  \text{softmax} (\frac{\mathbf{p}_i^j}{\tau_s}) \right).
\end{equation*}

Following~\cite{dinov2}, we further include the KoLeo regularizer~\cite{koleo} to promote a more uniform feature distribution within each batch.
The total SSL objective is:
\begin{equation}
  \mathcal{L}_{\text{SSL}}
  = \mathcal{L}_{\text{distill}}
  + \lambda_\text{mask} \times \mathcal{L}_{\text{mask}}
  + \lambda_\text{koleo} \times \mathcal{L}_{\text{koleo}},
  \label{equ:ssl}
\end{equation}
where $\lambda_\text{mask}$=2 and $\lambda_\text{koleo}$=0.1 as in ~\cite{dinov2}.
By integrating $\mathcal{L}_{\text{SSL}}$ with $\mathcal{L}_{\text{VL}}$, the model learns representations that are both semantically rich and spatially grounded.

\subsection{Dense Structured Supervision} \label{sec:dense}

While VL and SSL provide semantic and structural regularities, they often lack the explicit spatial precision required for dense downstream tasks.
To bridge this gap, we incorporate dense structured supervision by distilling knowledge from high-capacity expert models into our shared encoder.
Concretely, we introduce two complementary objectives: (i) region-level grounding for object-level semantics and (ii) per-pixel depth regression for fine-grained geometry.

\vspace{5pt}
\noindent\textbf{Region-level Grounding.}
To enforce spatial grounding, we incorporate regional-textual correspondences.
For each image $v_i$, we sample up to four region–text pseudo-labels to balance supervision density and computational cost.
The visual and textual inputs are encoded into patch-level features $\mathbf{z}_i$ and region text embeddings ${\mathbf{t}_{i,j}}$, respectively.
Following CLOC~\cite{cloc}, we utilize the Soft Region Aggregation mechanism to handle potential misalignments in pseudo-labels.
Specifically, a lightweight Transformer layer processes the patch-level features $\mathbf{z}_i$, conditioned on the positional encodings derived from each bounding box. This learnable attention allows the model to adaptively integrate context beyond rigid box boundaries, mitigating noise in pseudo labels.
The resulting soft region embeddings $\mathbf{z}_{i,j}$ are aligned with their corresponding text embeddings ${\mathbf{t}_{i,j}}$ using the sigmoid contrastive loss (Eq.~\eqref{equ:vl}), yielding the grounding loss $\mathcal{L}_\text{ground}$.

\vspace{5pt}
\noindent\textbf{Pixel-level Geometric Supervision.}
To capture dense geometric priors, we incorporate monocular depth estimation supervised by high-fidelity pseudo labels.
We attach a lightweight DPT head~\cite{dpt} to the shared visual encoder, which aggregates multi-layer features $\{\mathbf{z}_i^{(l)}\}_{l \in \mathcal{S}}$ from selected ViT layers to predict a dense depth map $\hat{d}_i \in \mathbb{R}^{H \times W}$.
The prediction is supervised using a combination of a scale- and shift-invariant loss alongside a multi-scale gradient loss from MiDaS~\cite{birkl2023midas}, jointly denoted as $\mathcal{L}_\text{depth}$.
To mitigate the influence of potential artifacts in pseudo labels, we adopt a top-$K$ denoising strategy, discarding the highest 10\% of per-pixel losses within each sample during training.
This encourages the encoder to prioritize high-confidence geometric structures, providing a structural anchor that complements the global semantic objectives.
Detailed loss formulations are provided in the supplementary material~\ref{sec:additional_training}.

\vspace{5pt}
By unifying region-level semantics and pixel-level geometry, the total dense supervision loss is defined as:
\begin{equation}
  \mathcal{L}_{\text{Dense}}
  = \mathcal{L}_{\text{ground}}
  + \mathcal{L}_{\text{depth}}.
  \label{equ:st}
\end{equation}

\subsection{Joint Optimization}
We optimize the shared visual encoder by jointly minimizing a multi-task loss that integrates the VL, SSL, and dense signals introduced in Eqs.~\eqref{equ:vl}--\eqref{equ:st}:
\begin{equation*}
  \mathcal{L}_{\text{total}} =
  \mathcal{L}_{\text{VL}}
  +  \mathcal{L}_{\text{SSL}}
  +  \mathcal{L}_{\text{Dense}}.
\end{equation*}
We provide a detailed gradient analysis in the supplementary material~\ref{sec:gradient_analysis}.
Empirically, we observe that uniform loss weighting yields stable convergence and competitive performance. 
As the constituent objectives naturally align in both magnitude (1--10) and gradient scale, the optimization process is not dominated by any individual task.
Consequently, our framework remains robust without requiring exhaustive hyperparameter tuning.
Furthermore, since every training image is fully equipped with all task annotations, we circumvent the need for complex task-specific sampling or data balancing. This unified training interface enables the shared encoder to efficiently internalize a comprehensive spectrum of visual information within a single forward-backward pass.

%% file: images/framework.tex
\begin{figure*}[t]
  \centerline{\includegraphics[width=\linewidth]{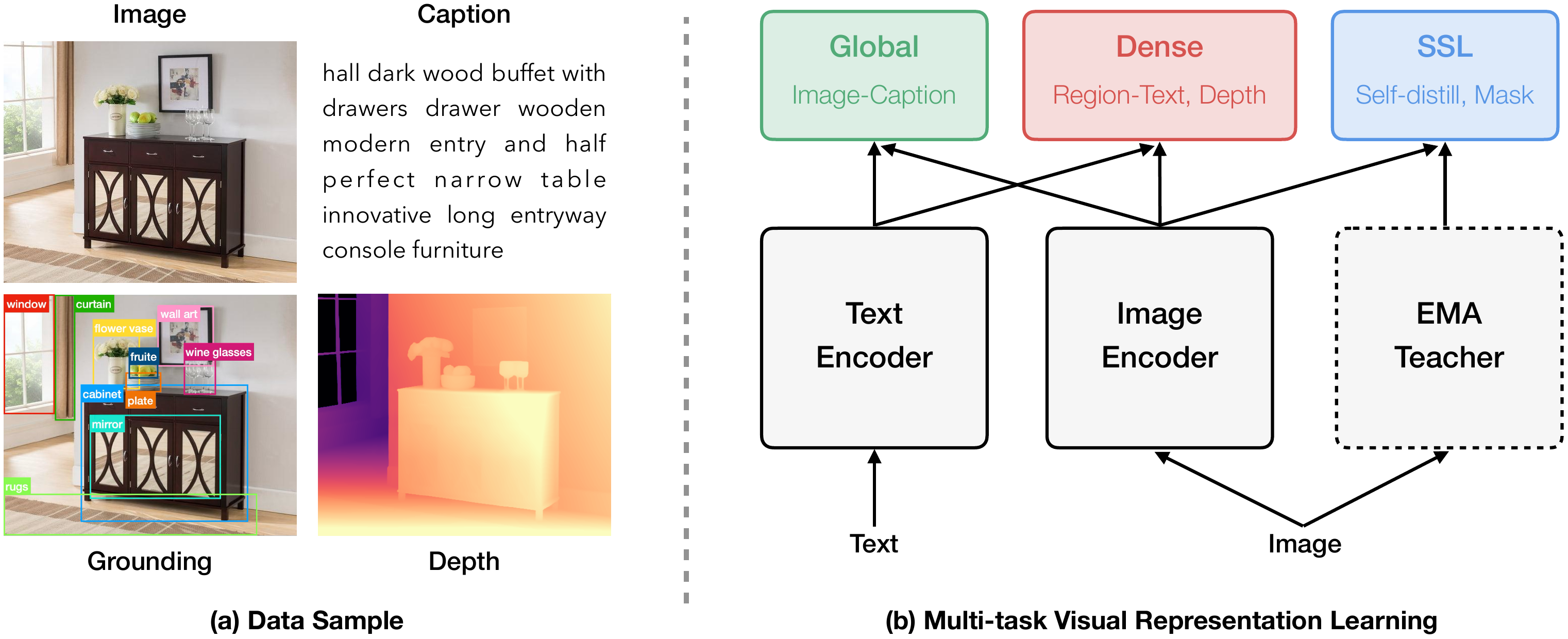}}
  \vspace{-5pt}
  \caption{
    \textbf{Overview of our \ourmethod~framework.}
    \textbf{(a)} Each image is paired with a web-crawled caption, and augmented with pseudo region–text pairs and relative depth maps generated by teacher models.
    \textbf{(b)} \ourmethod~jointly learns from three complementary supervision types:
    \textcolor[RGB]{60,171,120}{Global (image–caption contrast)},
    \textcolor[RGB]{214,43,79}{Dense (region–text alignment, depth)},
    and \textcolor[RGB]{88,130,230}{SSL (self-distillation, masked feature prediction)}.
    A shared image encoder is optimized together with a text encoder and an EMA teacher.
  }\label{fig:framework}
  \vspace{-10pt}
\end{figure*}

%% file: sections/4_experiments.tex
\section{Experiments}

\input{tables/compute}

\subsection{Training Details}
\label{sec:training_details}

\noindent\textbf{Architecture.} We evaluate our \ourmethod~framework using \texttt{ViT-B/16-224} and \\ \texttt{ViT-L/16-256} backbones, following the SigLIP~\cite{siglip} architecture and its text tokenizer.
To accommodate the multi-task nature of \ourmethod, we attach lightweight heads for SSL (MLP heads), grounding (Transformer aggregator), and depth estimation (DPT head).
Table~\ref{table:compute} summarizes the incremental cost of each task, while architectural specifications are detailed in the supplementary material.

\vspace{5pt}
\noindent\textbf{Training Data Curation.}
All experiments are conducted on the DataComp dataset~\cite{datacomp}.
We crawled 200 million raw image–text pairs and applied a filtering pipeline.
We discard images with a shorter side of $<224$ pixels and downsample high-resolution images to a maximum dimension of 1024 pixels while maintaining aspect ratios.
This filtering removes low-information outliers and ensures that the ``expert'' models—OWLv2 and Depth Anything V2—operate within their optimal receptive fields.
We then apply the pseudo-labeling procedure described in Section~\ref{sec:formulation}, resulting in 100 million ``fully-labeled'' samples, each containing a caption, region–text pairs, and a relative depth map. More data curation details can be found in the supplementary material~\ref{sec:data_curation}.

\vspace{5pt}
\noindent\textbf{Optimization Configurations.}
We pretrain our models from scratch, and investigate the scaling behavior of \ourmethod~at three data scales: 10M, 50M, and 100M samples.
The 10M and 50M scales are used for ablation and are trained for 20 and 10 epochs, respectively, with a global batch size of 4k.
Our full-scale 100M model is trained for 32 epochs with a global batch size of 8k, following the CLIP~\cite{clip} protocol for fair comparison.
We use the Adam optimizer with a learning rate of $5\times10^{-4}$, and a linear warmup over the first 1k steps.
Training uses BF16 mixed precision for efficiency, except for the DPT depth head, which is kept in FP32 to ensure numerical stability when regressing continuous geometric values. Experiments are conducted on 8-64 NVIDIA H200 GPUs. The final \texttt{MTV-B} model takes 4.4k GPU hours, and \texttt{MTV-L} costs 14.4k GPU hours.

\subsection{Evaluation Details}

We evaluate the learned representations across a broad range of downstream tasks, spanning global semantic understanding and fine-grained spatial reasoning. As summarized in Table~\ref{table:benchmark}, each task is assessed on one or more representative benchmarks. Below, we outline the setup for each task.

\input{tables/benchmark}

\vspace{5pt}
\noindent\textbf{Zero-shot Image Classification.}
We report Top-1 accuracy on the ImageNet-1k~\cite{imagenet} validation set using the standard prompt-based zero-shot protocol.

\vspace{5pt}
\noindent\textbf{Zero-shot Retrieval.}
For cross-modal alignment, we evaluate image-to-text (I$\to$T) and text-to-image retrieval (T$\to$I) on COCO~\cite{coco}, reporting Recall@1.

\vspace{5pt}
\noindent\textbf{MLLM VQA.}
We integrate the frozen visual encoder into the TinyLLaVA framework~\cite{tinyllava} with a \texttt{Qwen2.5-3B} LLM~\cite{qwen2_5}.
Following the standard two-stage recipe, we train the model on LLaVA-558K, followed by supervised fine-tuning on LLaVA-NEXT-779K~\cite{llavanext}.
Evaluation is conducted on three vision-centric VQA benchmarks categorized by~\cite{tong2024cambrian}, including MMVP~\cite{mmvp}, CVBench~\cite{tong2024cambrian}, and RealWorldQA~\cite{realworldqa}. We report the average score across these benchmarks.

\vspace{5pt}
\noindent\textbf{Semantic Segmentation.}
We evaluate on ADE20k~\cite{ade20k} using a frozen visual encoder and a linear classification head to predict per-patch semantic labels.
Training runs for 40k steps with batch size 32, learning rate 4e-5, and input resolution $576\times576$.
Performance is measured by mIoU.

\noindent\textbf{Geometric and Semantic Correspondence.}
To investigate the model's 3D awareness, we utilize the Probe3D framework~\cite{probe3d}.
We assess \textit{geometric correspondence} on NAVI~\cite{navi}, which requires matching pixels originating from the same 3D point across different viewpoints.
We further evaluate \textit{semantic correspondence} on SPair~\cite{spair}, testing the alignment of semantically similar parts across diverse instances.
Both tasks are measured using Average Recall across various viewpoints or distances.

\vspace{5pt}
\noindent\textbf{Monocular Depth Estimation.}
Following~\cite{dinov2,siglip2,fit3d}, we concatenate the global token to each patch token, and train a linear head to predict depth across 256 discretized bins. A classification loss is used during training on NYUv2~\cite{nyu}, with evaluation on its test set. See~\cite{fit3d} for more details.

\subsection{Ablations and Analysis}

We organize our analysis around four core questions:

\vspace{5pt}
\noindent\textbf{Q1: Individual Task Contributions.}

\input{tables/task_ablate}

We first break down the individual contribution of each task to isolate the sources of the gains. Table~\ref{table:task_ablate} reveals three observations: \textit{(i) Universal improvement}: adding any single objective on top of VL pretraining yields measurable gains on every benchmark, indicating that the auxiliary signals provide non-redundant supervision rather than task-specific overfitting. Surprisingly, the depth supervision is beneficial to semantic tasks ({\em e.g.}, +2.8 IN-1k and +2.5 COCO$^{\text{I$\to$T}}$ for ViT-L at 50M), even though the depth signal itself carries no semantic information.
We conjecture that regressing per-pixel depth forces the encoder to delineate
object boundaries, yielding more object-centric
features that in turn benefit instance-level recognition and alignment.
\textit{(ii) Specialized gains}: each objective excels on tasks aligned with its inductive bias. SSL primarily boosts semantic recognition and alignment (IN-1k, COCO, ADE20k, SPair), Depth contributes vital geometric priors (NAVI, NYUv2), and Grounding serves as a critical bridge that benefits both high-level semantics and fine-grained spatial structures. \textit{(iii) Cross-task spillover}: despite these specializations, we observe meaningful improvements outside each task's primary domain (e.g., Depth modestly helping semantic understanding), suggesting that the shared encoder is learning transferable primitives rather than siloed features. 

Taken together, the ablations indicate that \textbf{each objective contributes a distinct yet complementary signal}, laying the foundation for the monotonic gains observed in the subsequent multi-task expansion of Q2.
\vspace{5pt}

\noindent\textbf{Q2: Marginal Task Gains.}

We then examine whether integrating heterogeneous objectives is consistently beneficial. Table~\ref{table:task_scaling} presents the incremental performance along a principled task expansion path: $\text{VL} \to +\text{SSL} \to +\text{Ground.} \to +\text{Depth}$.

\input{tables/task_scaling}

Across data and model scales, performance almost always improves monotonically at every step. \Plusvalue{Absolute Gain $\Delta$} highlights that the full multi-task framework is strictly better than the VL-only baseline.
This demonstrates that \textbf{rather than interference, these objectives provide complementary supervision that enriches the shared representation.}

We also observe distinct roles for each objective:
(i) SSL acts as a powerful general-purpose enhancer, significantly lifting both semantic ($+14.4\%$ on IN-1k for \texttt{ViT-L/10M}) and spatial benchmarks (+10.4 on ADE20k). This suggests that local structural regularities provide a necessary geometric scaffold for global semantics.
(ii) Since grounding serves as a fine-grained vision-language signal, it not only boosts spatial tasks but also provides benefits to global multimodal understanding, such as the $+3.8$ R@1 gain in COCO I$\to$T retrieval.
(iii) Adding depth supervision consistently provides the final refinement, yielding maximum gains of 0.9 on
IN-1k, 2.2 on ADE20k , and 0.051 on NYUv2, {\em etc.}

While the gains exhibit diminishing marginal returns as the representation nears saturation, the large initial jump from SSL should be interpreted with caution: this is a consequence of the chosen expansion order and the inherent spatial weakness of the VL-only baseline. Nevertheless, the consistent positive deltas in Table~\ref{table:task_ablate} and \ref{table:task_scaling} verify the robustness of the joint optimization.

\noindent\textbf{Q3: Synergy {\em vs.} Interference among Objectives.}

We further investigate how heterogeneous objectives interact during joint optimization.
To quantify their complementarity, we define \textit{Task Synergy} as:
\begin{equation}
  \text{Synergy}(A, B) = \frac{\Delta AB - \max(\Delta A, \Delta B)} {\max(\Delta A, \Delta B)},
  \label{equ:synergy}
\end{equation}
where $\Delta A$ and $\Delta B$ denote the individual performance gains over the VL baseline, and $\Delta AB$ is the gain achieved by their combination.
Intuitively, a positive score indicates that the tasks are mutually beneficial, providing gains beyond those attainable by the superior individual task alone.

\input{tables/task_synergy}

As shown in Table~\ref{table:task_synergy}, all task pairs exhibit positive average synergy across data and model scales, typically ranging from $20\%$ to $50\%$. This empirically confirms that \textbf{the objectives are fundamentally complementary rather than interfering}.
Task pairs involving grounding show particularly robust synergy, reinforcing our argument that region–text alignment serves as a critical bridge between high-level semantics and fine-grained spatial structures.

We also observe distinct scaling trends between model and data scales: For the \texttt{B/16} model, increasing data from 10M to 50M samples leads to more stable and stronger synergy. While synergy remains positive for the larger \texttt{L/16} model, the relative scores are generally lower compared to \texttt{B/16}. This is expected as higher-capacity baselines naturally leave less relative headroom for additional supervision, even as absolute performance continues to rise.

\vspace{5pt}
\noindent\textbf{Q4: Scaling Effects across Data and Model Capacity.}

We finally investigate how \ourmethod~scales with data size and model capacity.
As illustrated in Fig.~\ref{fig:data_scaling}, scaling the data yields broadly monotonic improvements.
Notably, our \texttt{ViT-Base} model trained on only 100M samples consistently outperforms the \texttt{CLIP-Base} baseline, which was trained on 400M samples.
This demonstrates that \textbf{multi-task supervision can be more data-efficient than pure VL pretraining for building versatile visual encoders}.

\input{images/data_scaling}
\input{images/data_filtering}

However, the scaling behavior is not uniform across tasks. Geometric and semantic correspondence saturates and begins to decline once the number of seen samples exceeds 1B.
We explain this from three aspects:
\textit{(i) Task disparity}: correspondence tasks require multi-view consistency. While depth helps, it remains a monocular proxy. The observed saturation reflects the limits of {monocular} supervision for {multi-view} tasks. Future work may benefit from incorporating multi-view consistency supervision.
\textit{(ii) Data quality bottleneck}: DataComp contains many geometrically ``flat'' images that hinder geometric learning. Preliminary experiments on a 10M subset show that filtering by depth variance significantly alleviates saturation (Fig.~\ref{fig:data_filtering_effects}). More dedicated filtering methods could further improve performance, which we leave for future work.
\textit{(iii) Relative efficiency}: despite saturation, our model still outperforms CLIP/SigLIP on these tasks using much less training data (Table~\ref{table:sota}).

Moreover, model scaling from \texttt{B/16} to \texttt{L/16} (Table~\ref{table:task_ablate}, \ref{table:task_scaling}, \ref{table:sota}) improves performance across tasks under the same data scale, narrowing the gap to larger VL-only baselines while preserving strong spatial understanding.


\subsection{Comparison with Large-scale VLMs}

\input{tables/sota}

Previous ablations (Table~\ref{table:task_ablate}--\ref{table:task_synergy}) are strictly controlled under an identical data budget and training setup to ensure clear causal attribution.
Notably, our VL+SSL configuration serves as a representative proxy for several existing multi-task frameworks~\cite{slip,silc,tips}, while the further integration of dense supervision yields substantial gains beyond them.

Table~\ref{table:sota} compares \ourmethod~with established VLMs, including CLIP as a widely used reference with comparable data scale, and SigLIP / SigLIP2 as stronger baselines trained on significantly larger datasets.

On \textbf{global semantic tasks} such as classification, retrieval, and VQA, our models consistently outperform CLIP, despite using only one-quarter of its training data. As expected, \ourmethod~lag behind SigLIP/SigLIP2, which are trained on $100\times$ more data, but the gap remains modest given the vast scale difference.
On \textbf{fine-grained spatial tasks}, the advantages of \ourmethod~are most prominent. Ours tail SigLIP2 on semantic segmentation, while achieving clearly better performance on correspondence and depth estimation.

Taken together, these results demonstrate that multi-task training offers a compelling and scalable alternative to purely vision–language or self-supervised pretraining for general-purpose visual representation learning.

%% file: tables/compute.tex
\begin{table}[h]
  \vspace{-15pt}
  \setlength{\tabcolsep}{1.5mm}
  \centering

  \caption{
    \textbf{Training Cost} for a \texttt{ViT-B/16} model under different training task combinations.
    \addvalue{Numbers in parentheses} indicate the increase over the previous row.
    $^\star$The EMA teacher used in SSL is frozen during training.
  }\label{table:compute}
  \vspace{-5pt}

  \resizebox{.75\linewidth}{!}{
    \begin{tabu}{l ll}
      \toprule
      Training Tasks & Trainable Parameters & Training Time \\
      \midrule

      VL                         & 203.2 M & $1.0 \times$ \\
      VL + SSL$^\star$           & 232.6 M\addvalue{+29.4 M} & $1.2 \times$\addvalue{+20\%} \\
      VL + SSL$^\star$ + Ground.         & 239.7 M\addvalue{+7.1 M}  & $1.5 \times$\addvalue{+25\%} \\
      VL + SSL$^\star$ + Ground. + Depth & 250.3 M\addvalue{+10.6 M} & $1.7 \times$\addvalue{+13\%} \\

      \bottomrule
    \end{tabu}
  }
  \vspace{-25pt}
\end{table}

%% file: tables/benchmark.tex
\begin{wraptable}[12]{R}{0.6\textwidth}
  \vspace{-30pt}
  \setlength{\tabcolsep}{1.5mm}
  \centering

  \caption{
    \textbf{Summary of evaluation benchmarks.}
  }
  \label{table:benchmark}
  \vspace{5pt}

  \resizebox{0.95\linewidth}{!}{
    \begin{tabu}{lll}
      \toprule
      Task & Metric & Benchmark \\
      \midrule
      \rowcolor{gray!10} \textit{\textcolor{black!60}{Global Semantic Understanding}} && \\
      \hspace{0.5em} Zero-shot Classification & Top-1 Acc. & ImageNet-1k \\
      \hspace{0.5em} Zero-shot Retrieval & Recall@1 & COCO \\
      \addlinespace
      \hspace{0.5em} \multirow{3}{*}{Visual Question-Answering} & \multirow{3}{*}{Score}
      & MMVP, \\
      && CVBench, \\
      && RealWorldQA \\

      \midrule

      \rowcolor{gray!10} \textit{\textcolor{black!60}{Fine-grained Spatial Reasoning}} && \\
      \hspace{0.5em} Semantic Segmentation & mIoU  & ADE20k \\
      \hspace{0.5em} Geometric Correspondence   & Avg. Recall & NAVI \\
      \hspace{0.5em} Semantic Correspondence  & Avg. Recall & SPair \\
      \hspace{0.5em} Monocular Depth  & RMSE & NYUv2 \\
      \bottomrule
  \end{tabu}}
\end{wraptable}

%% file: tables/task_ablate.tex
\begin{table}[t]
  \setlength{\tabcolsep}{1.5mm}
  \centering
  \caption{
    \textbf{Individual Task Contributions} (50M Data).
    10M results are omitted due to space constraints, but show similar trends.
  }\label{table:task_ablate}
  \vspace{-5pt}

  \resizebox{\linewidth}{!}{
    \begin{tabu}{ll llllll}
      \toprule
      ViT & Training Tasks & IN-1k & COCO$_\text{I}$ & ADE20k & NAVI & SPair & NYUv2 $\downarrow$ \\
      \midrule


      \multirow{4}{*}{\texttt{B-16}}
      & VL & 55.7 & 43.0 & 32.2 & 39.8 & 18.0 & 0.604 \\
      & VL + SSL & \textbf{61.4\plusvalue{+5.7}} & \textbf{48.0\plusvalue{+5.0}} & \textbf{39.9\plusvalue{+7.7}} & 40.4\plusvalue{+0.6} & \textbf{22.7\plusvalue{+4.7}} & 0.556\plusvalue{-.048} \\
      & VL + Ground. &  59.5\plusvalue{+3.8} & 46.8\plusvalue{+3.8} & 37.6\plusvalue{+5.4} & 41.8\plusvalue{+2.0} & 21.2\plusvalue{+3.2} & 0.541\plusvalue{-.063} \\
      & VL + Depth &  57.7\plusvalue{+2.0} & 44.0\plusvalue{+1.0} & 36.0\plusvalue{+3.8} & \textbf{42.2\plusvalue{+2.4}} & 21.3\plusvalue{+3.3} & \textbf{0.511\plusvalue{-.093}} \\
      \midrule


      \multirow{4}{*}{\texttt{L/16}}
      & VL & 60.8 & 46.3 & 37.3 & 39.9 & 19.8 & 0.575 \\
      & VL + SSL & \textbf{65.4\plusvalue{+4.6}} & 48.0\plusvalue{+1.7} & \textbf{45.5\plusvalue{+8.2}} & 43.2\plusvalue{+3.3} & \textbf{27.3\plusvalue{+7.5}} & 0.501\plusvalue{-.074} \\
      & VL + Ground. &  64.1\plusvalue{+3.3} & \textbf{50.5\plusvalue{+4.2}} & 42.1\plusvalue{+4.8} & 43.3\plusvalue{+3.4} & 25.4\plusvalue{+5.6} & 0.508\plusvalue{-.067} \\
      & VL + Depth &  63.6\plusvalue{+2.8} & 48.8\plusvalue{+2.5} & 41.3\plusvalue{+4.0} & \textbf{43.6\plusvalue{+3.7}} & 22.7\plusvalue{+2.9} & \textbf{0.467\plusvalue{-.108}}  \\

      \bottomrule
  \end{tabu}}
\end{table}

%% file: tables/task_scaling.tex
\begin{table}[t]
  \centering
  \caption{
    \textbf{Marginal Task Gains} across data and model scales.
    \plusvalue{Numbers in parentheses} indicate the incremental increase over the previous row, while \Plusvalue{Absolute Gain $\Delta$} denotes the overall improvement of the full multi-task model over the VL-only baseline.
  }\label{table:task_scaling}
  \vspace{-5pt}

  \resizebox{\linewidth}{!}{
    \begin{tabu}{lll c llll c llll}
      \toprule
      \multirow{2}{*}{ViT} & \multirow{2}{*}{Data} & \multirow{2}{*}{Training Tasks} && IN-1k & \multicolumn{2}{c}{COCO} & VQA && ADE20k & NAVI & SPair & NYUv2 \\
      \cmidrule{5-8} \cmidrule{10-13}
      &&&& Acc. & T$\to$I & I$\to$T & Score && mIoU & Recall & Recall & RMSE $\downarrow$ \\
      \midrule

      \multirow{5}{*}{\texttt{B/16}} & \multirow{5}{*}{10M}
      & VL && 36.2 & 14.8 & 21.9 & 41.1 && 27.5 & 39.5 & 17.0 & 0.643 \\
      && VL + SSL && 43.7\plusvalue{+7.5} & 19.7\plusvalue{+4.9} & 28.6\plusvalue{+6.7} & 43.1\plusvalue{+2.0} && 36.2\plusvalue{+8.7} & 41.5\plusvalue{+2.0} & 21.1\plusvalue{+4.1} & 0.568\plusvalue{-.075} \\
      && VL + SSL + Ground. && 49.0\plusvalue{+5.3} & 23.5\plusvalue{+3.8} & 33.9\plusvalue{+5.3} & 43.2\plusvalue{+0.1} && 39.5\plusvalue{+3.3} & 43.3\plusvalue{+1.8} & 22.4\plusvalue{+1.3} & 0.537\plusvalue{-.031} \\
      && VL + SSL + Ground. + Depth && 49.7\plusvalue{+0.7} & 23.9\plusvalue{+0.4} & 34.1\plusvalue{+0.2} & 43.4\plusvalue{+0.2} && 41.7\plusvalue{+2.2} & 43.6\plusvalue{+0.3} & 22.8\plusvalue{+0.4} & 0.512\plusvalue{-.025} \\
      \rowcolor{green!10}
      && \Plusvalue{Absolute Gain $\Delta$} && \Plusvalue{+13.5} & \Plusvalue{+9.1} & \Plusvalue{+12.2} & \Plusvalue{+2.3} && \Plusvalue{+14.2} & \Plusvalue{+4.1} & \Plusvalue{+5.8} & \Plusvalue{-0.131} \\

      \midrule

      \multirow{5}{*}{\texttt{B/16}} & \multirow{5}{*}{50M}
      & VL && 55.7 & 28.2 & 43.0 & 42.2 && 32.2 & 39.8 & 18.0 & 0.604 \\
      && VL + SSL && 61.4\plusvalue{+5.7} & 32.4\plusvalue{+4.2} & 48.0\plusvalue{+5.0} & 44.7\plusvalue{+2.5} && 39.9\plusvalue{+7.7} & 40.4\plusvalue{+0.6} & 22.7\plusvalue{+4.7} & 0.556\plusvalue{-.048} \\
      && VL + SSL + Ground. && 62.9\plusvalue{+1.5} & 34.7\plusvalue{+2.3} & 50.7\plusvalue{+2.7} & 45.1\plusvalue{+0.4} && 43.0\plusvalue{+3.1} & 43.8\plusvalue{+3.4} & 25.4\plusvalue{+2.7} & 0.513\plusvalue{-.043} \\
      && VL + SSL + Ground. + Depth && 62.9 & 35.0\plusvalue{+0.3} & 50.5\minusvalue{-0.2} & 45.8\plusvalue{+0.7} && 43.7\plusvalue{+0.7} & 44.1\plusvalue{+0.3} & 24.5\minusvalue{-0.9} & 0.482\plusvalue{-.031} \\
      \rowcolor{green!10}
      && \Plusvalue{Absolute Gain $\Delta$} && \Plusvalue{+7.2} & \Plusvalue{+6.8} & \Plusvalue{+7.5} & \Plusvalue{+3.6} && \Plusvalue{+11.5} & \Plusvalue{+4.3} & \Plusvalue{+6.5} & \Plusvalue{-0.122} \\

      \midrule

      \multirow{5}{*}{\texttt{L/16}} & \multirow{5}{*}{10M}
      & VL && 35.0 & 14.9 & 21.6 & 39.0 && 30.0 & 40.1 & 18.1 & 0.630 \\
      && VL + SSL && 49.4\plusvalue{+14.4} & 23.7\plusvalue{+8.8} & 34.0\plusvalue{+12.4} & 42.1\plusvalue{+3.1} && 40.4\plusvalue{+10.4} & 43.4\plusvalue{+3.3} & 25.6\plusvalue{+7.5} & 0.535\plusvalue{-.095} \\
      && VL + SSL + Ground. && 53.4\plusvalue{+4.0} & 25.6\plusvalue{+1.9} & 37.8\plusvalue{+3.8} & 43.7\plusvalue{+1.6} && 43.6\plusvalue{+3.2} & 44.3\plusvalue{+0.9} & 27.0\plusvalue{+1.4} & 0.498\plusvalue{-.037} \\
      && VL + SSL + Ground. + Depth && 54.3\plusvalue{+0.9} & 26.6\plusvalue{+1.0} & 38.6\plusvalue{+0.8} & 44.3\plusvalue{+0.6} && 45.6\plusvalue{+2.0} & 46.2\plusvalue{+1.9} & 27.5\plusvalue{+0.5} & 0.455\plusvalue{-.043} \\
      \rowcolor{green!10}
      && \Plusvalue{Absolute Gain $\Delta$} && \Plusvalue{+19.3} & \Plusvalue{+11.7} & \Plusvalue{+17.0} & \Plusvalue{+5.3} && \Plusvalue{+15.6} & \Plusvalue{+6.1} & \Plusvalue{+9.4} & \Plusvalue{-0.175} \\

      \midrule

      \multirow{5}{*}{\texttt{L/16}}  & \multirow{5}{*}{50M}
      & VL && 60.8 & 31.1 & 46.3 & 43.3 && 37.3 & 39.9 & 19.8 & 0.575 \\
      && VL + SSL && 65.4\plusvalue{+4.6} & 34.5\plusvalue{+3.4} & 48.0\plusvalue{+1.7} & 44.9\plusvalue{+1.6} && 45.5\plusvalue{+8.2} & 43.2\plusvalue{+3.3} & 27.3\plusvalue{+7.5} & 0.501\plusvalue{-.074} \\
      && VL + SSL + Ground. && 67.6\plusvalue{+2.2} & 37.4\plusvalue{+2.9} & 51.5\plusvalue{+3.5} & 45.9\plusvalue{+1.0} && 47.2\plusvalue{+1.7} & 43.7\plusvalue{+0.5} & 28.1\plusvalue{+0.8} & 0.472\plusvalue{-.029} \\
      && VL + SSL + Ground. + Depth && 67.9\plusvalue{+0.3} & 37.3\minusvalue{-0.1} & 52.3\plusvalue{+0.8} & 45.9 && 49.1\plusvalue{+1.9} & 45.9\plusvalue{+2.2} & 29.7\plusvalue{+1.6} & 0.421\plusvalue{-.051} \\
      \rowcolor{green!10}
      && \Plusvalue{Absolute Gain $\Delta$} && \Plusvalue{+7.1} & \Plusvalue{+6.2} & \Plusvalue{+6.0} & \Plusvalue{+2.6} && \Plusvalue{+11.8} & \Plusvalue{+6.0} & \Plusvalue{+9.9} & \Plusvalue{-0.154} \\

      \bottomrule
  \end{tabu}}
  \vspace{-15pt}
\end{table}

%% file: tables/task_synergy.tex
\begin{table}[t]
  \setlength{\tabcolsep}{1.5mm}
  \centering

  \caption{
    \textbf{Task Synergy} (\%) computed as Eq.~\eqref{equ:synergy}, measuring the relative gain from combining two tasks beyond the larger of their individual improvements over the VL-only baseline.
    $^{\star}$COCO is short for COCO I$\to$T retrieval.
  }
  \label{table:task_synergy}
  \vspace{-5pt}

  \resizebox{\linewidth}{!}{
    \begin{tabu}{lcccc c rrrrrrr}
      \toprule
      \multirow{2}{*}{ViT} & \multirow{2}{*}{Data} & \multirow{2}{*}{Baseline} & \multirow{2}{*}{TaskA} & \multirow{2}{*}{TaskB} && \multicolumn{7}{c}{Synergy (\%) } \\
      \cmidrule{7-13}
      &&&&&& IN-1k & COCO$^{\star}$ & ADE20k & NAVI & SPair & NYUv2 & \textbf{Average} \\
      \midrule

      \multirow{3}{*}{\texttt{B/16}} & \multirow{3}{*}{10M} & \multirow{3}{*}{VL}
      & SSL & Ground.  && 62.0 & 64.4 & 37.9 & 72.7 & 31.7 & 40.8 & 51.7 \\
      &&& SSL & Depth && 50.7 & 41.8 & 31.0 & -10.0 & 26.8 & 56.6 & 32.9 \\
      &&& Depth & Ground. && 22.8 & 27.4 & 33.8 & 36.4 & 51.5 & 56.2 & 38.1 \\

      \midrule

      \multirow{3}{*}{\texttt{B/16}} & \multirow{3}{*}{50M} & \multirow{3}{*}{VL}
      & SSL & Ground.  && 26.3 & 54.0 & 40.3 & 100.0 & 57.4 & 44.4 & 53.7 \\
      &&& SSL & Depth && 8.8 & 4.0 & 27.3 & 75.0 & 61.7 & 29.0 & 34.3 \\
      &&& Depth & Ground. && 2.6 & 26.3 & 40.7 & 87.5 & 78.8 & 19.4 & 42.6 \\

      \midrule

      \multirow{3}{*}{\texttt{L/16}} & \multirow{3}{*}{10M} & \multirow{3}{*}{VL}
      & SSL & Ground.  && 27.8 & 30.6 & 30.8 & 27.3 & 18.7 & 38.9 & 29.0 \\
      &&& SSL & Depth && 17.4 & 21.8 & 27.9 & 24.2 & 24.0 & 53.6 & 28.1 \\
      &&& Depth & Ground. && 9.6 & 3.5 & 32.0 & 42.9 & 0.0 & 12.5 & 16.7 \\

      \midrule

      \multirow{3}{*}{\texttt{L/16}} & \multirow{3}{*}{50M} & \multirow{3}{*}{VL}
      & SSL & Ground.  && 47.8 & 33.8 & 30.7 & 21.8 & 10.7 & 39.2 & 30.7 \\
      &&& SSL & Depth && 43.5 & 92.0 & 19.5 & 9.1 & -34.7 & 4.6 & 22.3 \\
      &&& Depth & Ground. && 45.5 & 33.3 & 33.3 & 29.4 & 17.9 & 18.5 & 29.7 \\

      \bottomrule
  \end{tabu}}
  \vspace{-10pt}
\end{table}

%% file: images/data_scaling.tex
\begin{figure}[t]
  \centerline{\includegraphics[width=\linewidth]{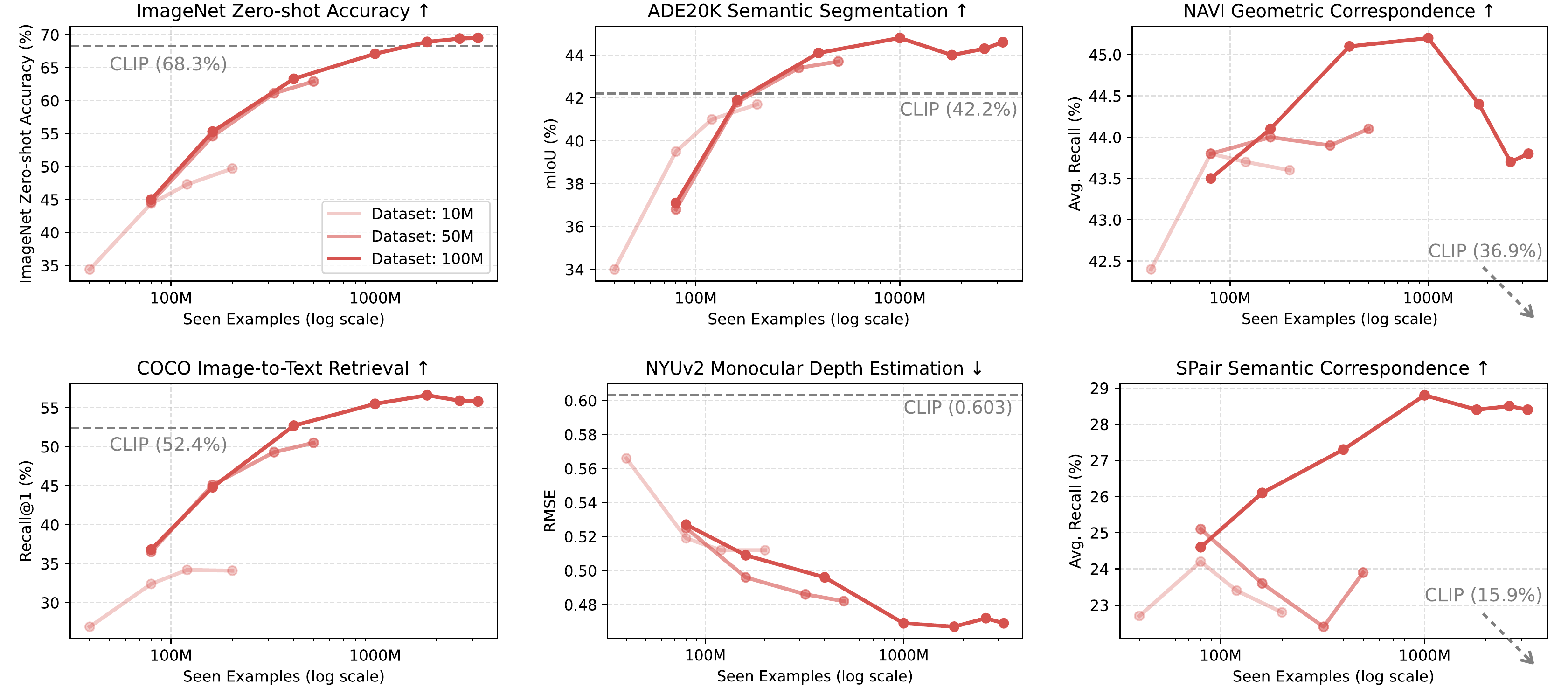}}
  \caption{
    \textbf{Data scaling behavior of multi-task visual learning.}
    \texttt{ViT-Base} models are trained with 10M, 50M, and 100M samples under our multi-task setting.
    \textcolor{gray}{\texttt{CLIP-Base}}~\cite{clip} is shown as gray dashed line.
    Lower$\downarrow$ is better for NYUv2; higher$\uparrow$ is better elsewhere.
  }\label{fig:data_scaling}
  \vspace{-15pt}
\end{figure}

%% file: images/data_filtering.tex
\begin{wrapfigure}[17]{R}{0.4\textwidth}
  \vspace{-20pt}
  \centering
  \includegraphics[width=0.4\textwidth]{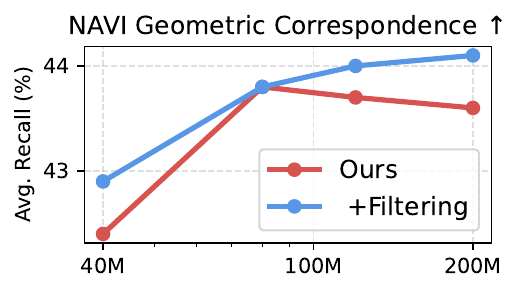}\\
  \includegraphics[width=0.4\textwidth]{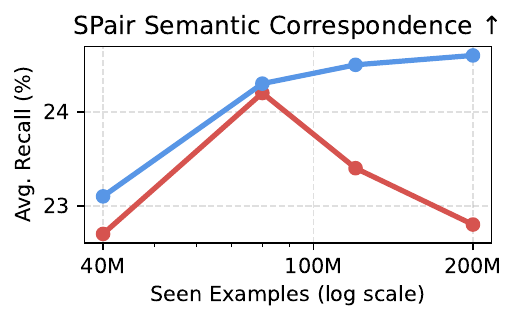}
  \vspace{-20pt}
  \caption{\textbf{Data filtering effects} (ViT-B, 10M data $\times$ 20 epochs).}
  \label{fig:data_filtering_effects}
\end{wrapfigure}

%% file: tables/sota.tex
\begin{table}[t]
  \setlength{\tabcolsep}{1.5mm}
  \centering

  \caption{
    \textbf{Comprehensive comparisons with large-scale VLMs.}
  }
  \label{table:sota}
  \vspace{-5pt}

  \resizebox{\linewidth}{!}{
    \begin{tabu}{llc c cccc c cccc}
      \toprule
      \multirow{2}{*}{ViT} & \multirow{2}{*}{Model} & \multirow{2}{*}{Data} && IN-1k & \multicolumn{2}{c}{COCO} & VQA && ADE20K & NAVI & SPair & NYUv2 \\
      \cmidrule{5-8} \cmidrule{10-13}
      &&&& Acc. & T$\to$I & I$\to$T & Score && mIoU & Recall & Recall & RMSE $\downarrow$\\
      \midrule

      \multirow{4}{*}{\texttt{B/16}} & CLIP~\cite{clip} & 400M && 68.3 & 33.1 & 52.4 & 44.8 && 42.2 & 36.9 & 15.9 & 0.603 \\
      & SigLIP~\cite{siglip} & 10B && 76.2 & 47.2 & 64.5 & 47.7 && 45.1 & 38.5 & 17.3 & 0.615 \\
      & SigLIP2~\cite{siglip2} & 10B && \textbf{78.2} & \textbf{52.1} & \textbf{68.9} & \textbf{48.1} && \textbf{46.0} & 38.0 & 20.8 & 0.562 \\
      \rowcolor{green!10}
      & \textbf{\ourmethod} & 100M && 69.5 & 41.2 & 57.1 & 45.6 && 44.6 & \textbf{43.8} & \textbf{28.4} & \textbf{0.469} \\

      \midrule

      \texttt{L/14}
      & CLIP~\cite{clip} & 400M && 75.5 & 36.5 & 56.3 & 46.5 && 46.2 & 36.0 & 20.5 & 0.588 \\
      \hdashline
      \multirow{3}{*}{\texttt{L/16}}
      & SigLIP~\cite{siglip} & 10B && 80.5 & 51.2 & 69.6 & 48.3 && 47.7 & 38.8 & 19.7 & 0.550 \\
      & SigLIP2~\cite{siglip2} & 10B && \textbf{82.5} & \textbf{54.7} & \textbf{71.5} & \textbf{52.5} && \textbf{51.6} & 43.2 & 29.9 & 0.484 \\
      \rowcolor{green!10}
      & \textbf{\ourmethod} & 100M && 75.2 & 43.0 & 58.1 & 47.6 && 48.2 & \textbf{45.0} & \textbf{30.1} & \textbf{0.414} \\

      \bottomrule
  \end{tabu}}
  \vspace{-5pt}
\end{table}

%% file: sections/5_conclusion.tex
\section{Conclusion}
In this work, we revisit large-scale multi-task training as a unified framework for learning general-purpose visual representations.
By jointly optimizing vision–language contrastive, self-supervised, and pseudo-labeled dense objectives, our model acquires strong semantic and geometric understanding within a single encoder.
Across extensive experiments, we demonstrate that multi-task supervision leads to steadily improving performance as more tasks are incorporated, exhibiting clear task synergy and favorable data- and model-scaling behavior.
These findings suggest that multi-task pretraining, fueled by high-quality pseudo-supervision, provides an effective and data-efficient route toward universal visual foundation models for broad visual perception and reasoning tasks.

%% file: sections/X_suppl.tex
\clearpage
\setcounter{page}{1}

\appendix

\section{Additional Implementation Details}

The main paper provides a high-level summary of our experimental setup while omitting certain lower-level details for clarity.
In this section, we offer a more comprehensive account of the implementation and training configurations.

\subsection{Data Curation} \label{sec:data_curation}

We adopt DataComp~\cite{datacomp} as the source for our pre-training data, which consists of 1.1 billion image–text pairs. Due to limitations in download bandwidth, we collected a subset of 200 million images from the full dataset. The download took around 40 days and resulted in nearly 21 TB of image–text data. We then applied the filtering strategy detailed in Section~\ref{sec:training_details}, obtaining a final corpus of 100 million image–text pairs used in our experiments.

We employ Depth Anything V2 (\texttt{DA2-Large})~\cite{depth_anythingv2} to compute a relative depth map for each image.
Our depth-processing pipeline attains a throughput of approximately 120K images per GPU-hour, so producing depth maps for 100M images on 64 GPUs takes about 13 hours.
This high throughput allows depth supervision to be easily scaled to much larger datasets.

For grounding supervision, we first use \texttt{RAM++}~\cite{ram++} to extract salient entity names, and then apply \texttt{OWLv2-Base}~\cite{owlv2} to localize the corresponding regions.
This pipeline can handle roughly 18K images per GPU-hour, leading to a total of 86 GPU-hours to produce region–text pairs for 100M images on 64 GPUs. While this is slower than depth estimation, the throughput is still high enough to feasibly generate pseudo-labels at the billion-sample scale.

\subsection{Additional Training Details}
\label{sec:additional_training}

\noindent\textbf{SSL Supervision.}
The projection heads for self-distillation and masked prediction follow the same architecture but do not share parameters.
Each head is a 3-layer MLP with L2 normalization, followed by a weight-normalized projection layer that maps to a 32k-dimensional prototype space.

The teacher network is updated as the exponential moving average (EMA) of the student using a fixed momentum of 0.994.
To avoid representation collapse, we apply both centering and sharpening after the projection head.
For centering, we maintain a running mean of student logits (momentum 0.9) and subtract this mean during training.
For sharpening, we use a fixed student temperature of $\tau_s = 0.1$ and linearly warm up the teacher temperature $\tau_t$ from 0.04 to 0.07.

\vspace{5pt}
\noindent\textbf{Grounding Supervision.}
As described in Section~\ref{sec:dense}, we follow the CLOC approach~\cite{cloc} to utilize the grounding supervision.
CLOC introduces a lightweight Transformer encoder, referred to as the \texttt{Prompter}, which extracts regional visual features conditioned on bounding box locations.
The \texttt{Prompter} takes three inputs:
(i) the visual encoder output $\mathbf{z}_i^*$ before final normalization and attentive pooling,
(ii) a location embedding $l_i$ derived from the bounding box coordinates (top-left and bottom-right), each encoded with positional embeddings, and
(iii) a learnable query token $\mathbf{o}$.
Given these inputs, the \texttt{Prompter} produces a regional visual feature $\textbf{o}_i = \texttt{Prompter}(\textbf{z}_i^*,\ l_i,\ \textbf{o})$, which is then aligned to the regional text feature through the same contrastive formulation used in Eq.~\eqref{equ:vl}.

\vspace{5pt}
\noindent\textbf{Depth Supervision.}
The depth head follows the DPT~\cite{dpt} decoder, a convolutional module that fuses features from four intermediate layers of the visual encoder:
$\{3,6,9,12\}$ for \texttt{ViT-B/16} and $\{6,12,18,24\}$ for \texttt{ViT-L/16}.

As described in Section~\ref{sec:dense}, we adopt the MiDaS objective~\cite{birkl2023midas}, combining a scale- and shift-invariant loss with a multi-scale gradient matching loss.

Regarding the scale- and shift-invariant loss, the pseudo depth $d$ and predicted depth $\hat{d}$ are first normalized by:
$$
d^* = \frac{d - t(d)}{s(d)}, \quad \hat{d}^* = \frac{\hat{d} - t(\hat{d})}{s(\hat{d})},
$$
where $t(d)=\text{median}(d), \quad s(d)=\frac{1}{HW}\sum_{j=1}^{HW} |d - t(d)|$.
The scale- and shift-invariant (ssi) loss is then defined as:
$$
\mathcal{L}_{\text{ssi}} = \frac{1}{2HW}\sum_{j=1}^{HW}|d_j^* - \hat{d}_j^*|.
$$
Following MiDaS~\cite{birkl2023midas}, we trim the 10\% largest residuals to mitigate pseudo-label noise: Let $r_j = |d_j^* - \hat{d}_j^*|$ sorted in ascending order, we keep only the smallest $U = 0.9HW$ residuals and define $\mathcal{L}_{\text{ssitrim}} = \frac{1}{2HW} \sum_{j=1}^{U} r_{j}$.

In addition, the gradient matching (gm) loss penalizes the depth discrepancies between the ground truth and the rescaled prediction across $K=4$ scales:
$$
\mathcal{L}_{\text{gm}} = \sum_{k=1}^K \sum_{j=1}^{HW} \left( | \nabla_x^k d_j - s\nabla_x^k \hat{d}_j | + | \nabla_y^k d_j - s\nabla_y^k \hat{d}_j | \right),
$$
where $\nabla_x^k$ and $\nabla_y^k$ denote finite differences along the horizontal and vertical directions at scale $k$.

The final depth loss is a weighted combination:
$$
\mathcal{L}_{\text{depth}} = \mathcal{L}_{\text{ssitrim}} + 2 \times \mathcal{L}_{\text{gm}}.
$$

\vspace{5pt}
\noindent\textbf{Computational Cost.}
For the largest 100M setting, our \texttt{\ourmethod-Base} is trained on 32 GPUs for 5.7 days, while \texttt{\ourmethod-Large} requires 64 GPUs for 9.4 days.

\section{Additional Experiments}

\subsection{Zero-shot Relative Depth Estimation}

\input{tables/sota_zs_depth}

Our model inherently supports zero-shot depth estimation through its pretraining phase. In Table~\ref{table:sota_zs_depth}, we evaluate \ourmethod~against state-of-the-art depth specialists on KITTI~\cite{kitti} and NYUv2~\cite{nyu}. Notably, despite being trained in a multi-task framework and operating at a significantly lower input resolution than dedicated models (e.g., Depth Anythings v1/v2 at $518^2$), \ourmethod~achieves competitive zero-shot performance across both benchmarks.


\input{images/visualization}

Beyond the quantitative metrics, Figure~\ref{fig:vis_depth} also presents qualitative comparisons. Our model generates depth maps that capture the correct global geometry and maintain a coherent scene structure across a wide variety of images. 
Relative to the depth-specialized DA2 model, some fine-scale details ({\em e.g.}, the cat’s whiskers or the chair back) appear less crisp, primarily due to the difference in resolution. 
Despite this, the predictions remain generally stable and accurate, indicating that low-resolution, pseudo-labeled depth supervision, when incorporated into multi-task pretraining, can still produce strong and well-generalized geometric representations.

\subsection{Ablation on the Depth Teacher} \label{sec:ablate_depth}
To evaluate our method's sensitivity to teacher capacity and the quality of pseudo-labels,
Table~\ref{table:depth_teacher} compares configurations using \texttt{Depth-Anything-V2-Small} (25M parameters, denoted as ``DA2-S''), \texttt{Depth-Anything-V2-Large} (335M parameters, ``DA2-L''), and a baseline without depth supervision (``N/A'').

Our analysis reveals several key insights:
\begin{itemize}
  \item \textbf{Stability in Semantic Tasks:}
  Performance on high-level semantic benchmarks remains remarkably stable as teacher capacity scales (e.g., IN-1k: 49.5 $\to$ 49.7; COCO$_{\text{T}\to\text{I}}$: 23.7 $\to$ 23.9; COCO$_{\text{I}\to\text{T}}$: 34.2 $\to$ 34.1).
  The marginal fluctuations suggest that distilling geometric priors from depth teachers primarily enriches structural understanding without compromising or significantly altering the semantic feature space.
  \item \textbf{Sensitivity of Spatial Understanding:}
  In contrast, tasks requiring fine-grained spatial reasoning exhibit a pronounced correlation with teacher capacity. Upgrading from DA2-S to DA2-L leads to consistent improvements across ADE20k (41.1 $\to$ 41.7), NAVI (42.1 $\to$ 43.6), SPair (21.7 $\to$ 22.8, and NYUv2 (0.527 $\to$ 0.512). This indicates that higher-fidelity pseudo-depth labels are crucial for capturing fine-grained structural details.
  \item \textbf{Efficacy of Structural Priors:}
  Notably, even the lightweight DA2-S teacher outperforms the ``N/A'' baseline across most benchmarks (IN-1k, COCO, ADE20k, and NYUv2). This suggests that the inclusion of depth as an auxiliary objective introduces beneficial structural priors, and these gains are largely orthogonal to the absolute precision of the teacher model itself.
\end{itemize}

\begin{table}[t]
  \setlength{\tabcolsep}{1.5mm}
  \centering
  \caption{
    \textbf{Ablation on Depth Teacher} with \texttt{ViT-B} and 10M Data.
  }\label{table:depth_teacher}
  \vspace{-5pt}
  \resizebox{.85\linewidth}{!}{
    \begin{tabu}{l ccccccc}
      \toprule
      Teacher & IN-1k & COCO$_{\text{T}\to\text{I}}$ & COCO$_{\text{I}\to\text{T}}$ & ADE20k & NAVI & SPair & NYUv2 $\downarrow$ \\
      \midrule
      N/A   & 49.0 & 23.5 & 33.9 & 39.5 & 43.3 & 22.4 & 0.537 \\
      DA2-S & 49.5 & 23.7 & \textbf{34.2} & 41.1 & 42.1 & 21.7 & 0.527 \\
      DA2-L & \textbf{49.7} & \textbf{23.9} & 34.1 & \textbf{41.7} & \textbf{43.6} & \textbf{22.8} & \textbf{0.512} \\
      \bottomrule
  \end{tabu}}
  \vspace{-10pt}
\end{table}

\subsection{Gradient Analysis} \label{sec:gradient_analysis}
We measure pairwise gradient cosine similarity on the shared ViT encoder (Table~\ref{tab:grad}). All pairs show positive alignment with moderate conflict rates and balanced norm ratios, confirming that no objective monopolizes the encoder.
These observations suggest limited optimization conflict: gradients are positively aligned on average, added objectives do not induce clear trade-offs on our benchmarks, and task pairs show positive empirical synergy.

\begin{table}[h]
\centering
\small
\setlength{\tabcolsep}{3.5pt}
\caption{\textbf{Gradient analysis.} Conflict: \% batches with negative cosine. Norm ratio: larger/smaller.}
\label{tab:grad}
\begin{tabular}{lccc}
\toprule
Task pair & Grad.\ cos.\ $\uparrow$ & Conflict $\downarrow$ & Norm ratio \\
\midrule
VL / SSL        & 0.18 & 21.4 & 1.23 \\
VL / Ground.    & 0.24 & 17.8 & 1.96 \\
SSL / Depth     & 0.14 & 24.1 & 1.84 \\
Ground. / Depth & 0.27 & 15.9 & 1.18 \\
\bottomrule
\end{tabular}
\end{table}

\subsection{Joint vs.\ curriculum learning.}
\begin{figure}[t]
\centerline{\includegraphics[width=.95\linewidth]{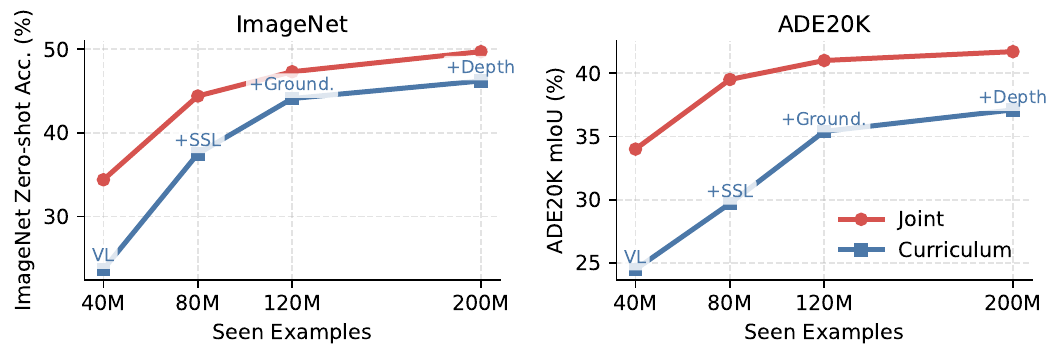}}
\vspace{-5pt}
\caption{\textbf{Joint vs.\ curriculum learning.}}\label{fig:curriculum}
\end{figure}

Under the same 10M data budget, we compare joint training with an additive VL$\rightarrow$SSL$\rightarrow$Grounding$\rightarrow$Depth curriculum.
Joint training uses all objectives from the start and consistently performs better (Fig.~\ref{fig:curriculum}), with final gains of $+3.5$ IN-1k and $+4.6$ ADE20K.
This suggests that early co-training better aligns semantic and spatial signals.

\section{Limitations and Future Work}

While our study provides a systematic examination of multi-task visual pretraining, several avenues for further investigation remain.


\vspace{5pt}
\noindent\textbf{More Dedicated Data Filtering.}
Our pseudo labels were generated automatically without exhaustive filtering. As a result, some images exhibit low-quality captions, noisy grounding annotations, or unreliable depth estimates. While our analysis suggests the model is robust to these imperfections, developing more sophisticated data-cleaning pipelines or noise-adaptive training objectives could further refine representation quality. Exploring the trade-off between "data quantity" and "label fidelity" in multi-task settings is a promising direction.

\vspace{5pt}
\noindent\textbf{Expanding the Modality Horizon.}
Our current framework focuses on single-view image pretraining.
Extending this multi-task formulation to temporal or multi-view data—such as video sequences or stereo imagery—could unlock richer supervision signals. Integrating objectives like optical flow, 4D consistency, or video-language alignment would be a natural progression toward learning truly universal and physically-grounded visual representations.

%% file: tables/sota_zs_depth.tex
\begin{wraptable}[12]{R}{0.48\textwidth}
  \centering
  \setlength{\tabcolsep}{1.5mm}
  \vspace{-10pt}
  \caption{
    \textbf{Zero-shot relative depth estimation,}
    measured by AbsRel $\downarrow$.
  }
  \vspace{5pt}
  \label{table:sota_zs_depth}
  \resizebox{0.96\linewidth}{!}{
    \begin{tabu}{lc cc}
      \toprule
      Model & Input & KITTI & NYUv2 \\
      \midrule
      Marigold~\cite{marigold} & $480\times640$ & 9.9 & 5.5 \\
      \midrule
      DAv1-B~\cite{depth_anything} & $518\times518$ & 8.0 & 4.6 \\
      DAv2-B~\cite{depth_anythingv2} & $518\times518$ & 7.8 & 4.9 \\
      \rowcolor{green!10}
      \ourmethod-B           & $224\times224$ & 8.9 & 6.0 \\
      \midrule
      MiDaS v3.1~\cite{birkl2023midas} & $518\times518$ & 12.7 & 4.8 \\
      DAv1-L~\cite{depth_anything}      & $518\times518$ & 7.6 & 4.3 \\
      DAv2-L~\cite{depth_anythingv2}      & $518\times518$ & 7.4 & 4.5 \\
      \rowcolor{green!10}
      \ourmethod-L                & $256\times256$ & 8.2 & 5.2 \\
      \bottomrule
  \end{tabu}}
\end{wraptable}

%% file: images/visualization.tex
\begin{figure}[!ht]
  \vspace{-5pt}
  \centerline{\includegraphics[width=\linewidth]{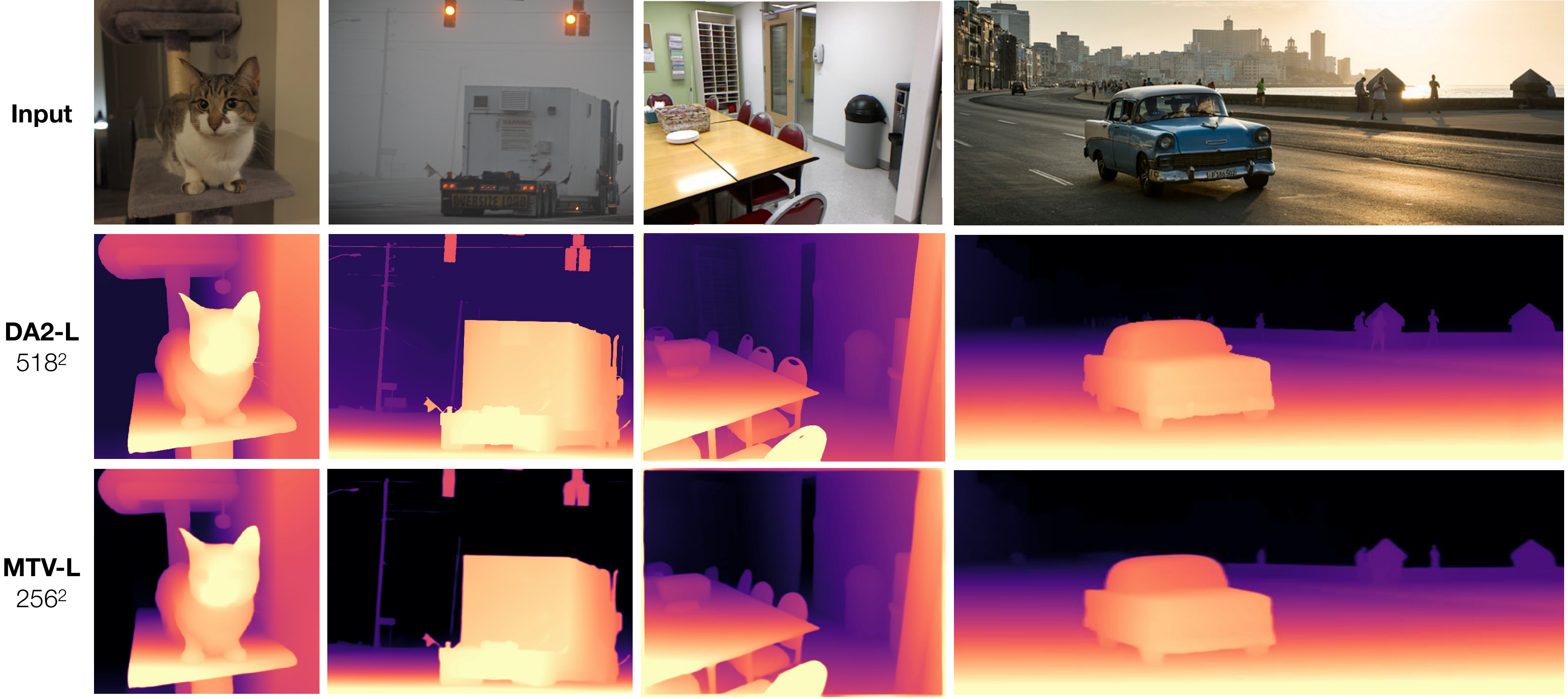}}
  \caption{
    \textbf{Visualizations of zero-shot relative depth estimation.}
  }\label{fig:vis_depth}
  \vspace{-15pt}
\end{figure}